\documentclass{article}

\usepackage[nonatbib,final]{midl_2018}

\usepackage{xcolor}
\definecolor{mydarkblue}{rgb}{0,0.15,0.7}
\usepackage[colorlinks, urlcolor=mydarkblue, citecolor=mydarkblue, linkcolor=mydarkblue]{hyperref}

\usepackage[utf8]{inputenc} % allow utf-8 input
\usepackage[T1]{fontenc}    % use 8-bit T1 fonts
\usepackage{hyperref}       % hyperlinks
\usepackage{url}            % simple URL typesetting
\usepackage{booktabs}       % professional-quality tables
\usepackage{amsfonts}       % blackboard math symbols
\usepackage{nicefrac}       % compact symbols for 1/2, etc.
\usepackage{microtype}      % microtypography
\usepackage{amsmath}
\usepackage{algorithmic}
\usepackage{algorithm}
\usepackage{graphicx}
\usepackage{subcaption}
\usepackage{siunitx}
\usepackage{textcomp}
\usepackage{amssymb}

\title{Detecting Lesion Bounding Ellipses With \\ Gaussian Proposal Networks}

\author{
  Yi Li \\
  Baidu Research Institute \\
  1195 Bordeaux Dr. Sunnyvale, CA 94089 \\
  \texttt{liyi17@baidu.com} \\
}

\begin{document}

\maketitle

\begin{abstract}
Lesions characterized by computed tomography (CT) scans, are arguably often 
elliptical objects. However, current lesion detection systems are predominantly
adopted from the popular Region Proposal Networks (RPNs)~\cite{ren2015faster} that only propose bounding
boxes without fully leveraging the elliptical geometry of lesions. In this paper,
we present Gaussian Proposal Networks (GPNs), a novel extension to RPNs, to detect
lesion bounding ellipses. Instead of directly regressing the rotation angle
of the ellipse as the common practice, GPN represents bounding ellipses as 2D
Gaussian distributions on the image plain and minimizes the Kullback-Leibler (KL)
divergence between the proposed Gaussian and the ground truth Gaussian for object
localization. We show the KL divergence loss approximately incarnates the regression
loss in the RPN framework when the rotation angle is 0. Experiments on the
DeepLesion~\cite{yan2018deeplesion} dataset show that GPN significantly outperforms
RPN for lesion bounding ellipse detection thanks to lower localization error.
GPN is open sourced at \url{https://github.com/baidu-research/GPN}.

\end{abstract}

\section{Introduction}
Current state-of-the-art object detection systems are predominantly based on deep
neural networks that learn to propose object regions which are usually
represented by bounding boxes~\cite{ren2015faster, liu2016ssd, lin2017feature, dai2016r, he2017mask, fu2017dssd}. 
Region Proposal Networks (RPNs), first introduced in Faster R-CNN~\cite{ren2015faster},
simultaneously predicts the objectness and regions bounds at every predefined anchor
location on a grid of feature map. When object regions are annotated as bounding
boxes, the region bounds in RPN are defined by the two center coordinates, the
width and the height of the object region with respect to the corresponding anchor.
In this case, regressing the regions bounds is directly optimizing the overlap
between the proposed bounding box and the ground truth bounding box.

Lesions characterized by computed tomography (CT) scans are often elliptical
and additional geometry information about the lesion regions may be annotated besides bounding boxes.
For example, the large-scale medical imaging dataset DeepLesion~\cite{yan2018deeplesion}
recently released from NIH is annotated with the response evaluation criteria in
solid tumors (RECIST) diameters. Each RECIST-diameter annotation consists of two
axies, the first one measures the longest diameter of the lesion and the second
one measures the longest diameter perpendicular to the first axis. Therefore,
the RECIST diameters closely represent the major and minor axises of a bounding
ellipse of the lesion.

Extensions for bounding ellipse detection based on the
RPN framework have been introduced to also predict the rotation angle of the
object~\cite{hu2017finding, shi2018real, liu2017learning, li2015convolutional, yang2016multi, mathias2014face, wang2016use, zhang2017s, opitz2016grid}.
However, due to the rotation angle, it is not trivial to directly optimizing the
overlap between the proposed bounding ellipse and the ground truth bounding
ellipse within the RPN framework. Most of the existing methods either directly
use an ellipse regressor to minimize the difference between the proposed angle
and the ground truth angle, e.g. in the co/tangent domain~\cite{hu2017finding, liu2017learning},
as an additional term in the regression loss for localization. Other methods
discretize the rotation angle first, and then predict the angle category as a
classification problem~\cite{shi2018real}.

\begin{figure}[t]
\begin{center}
   \includegraphics[width=0.70\linewidth]{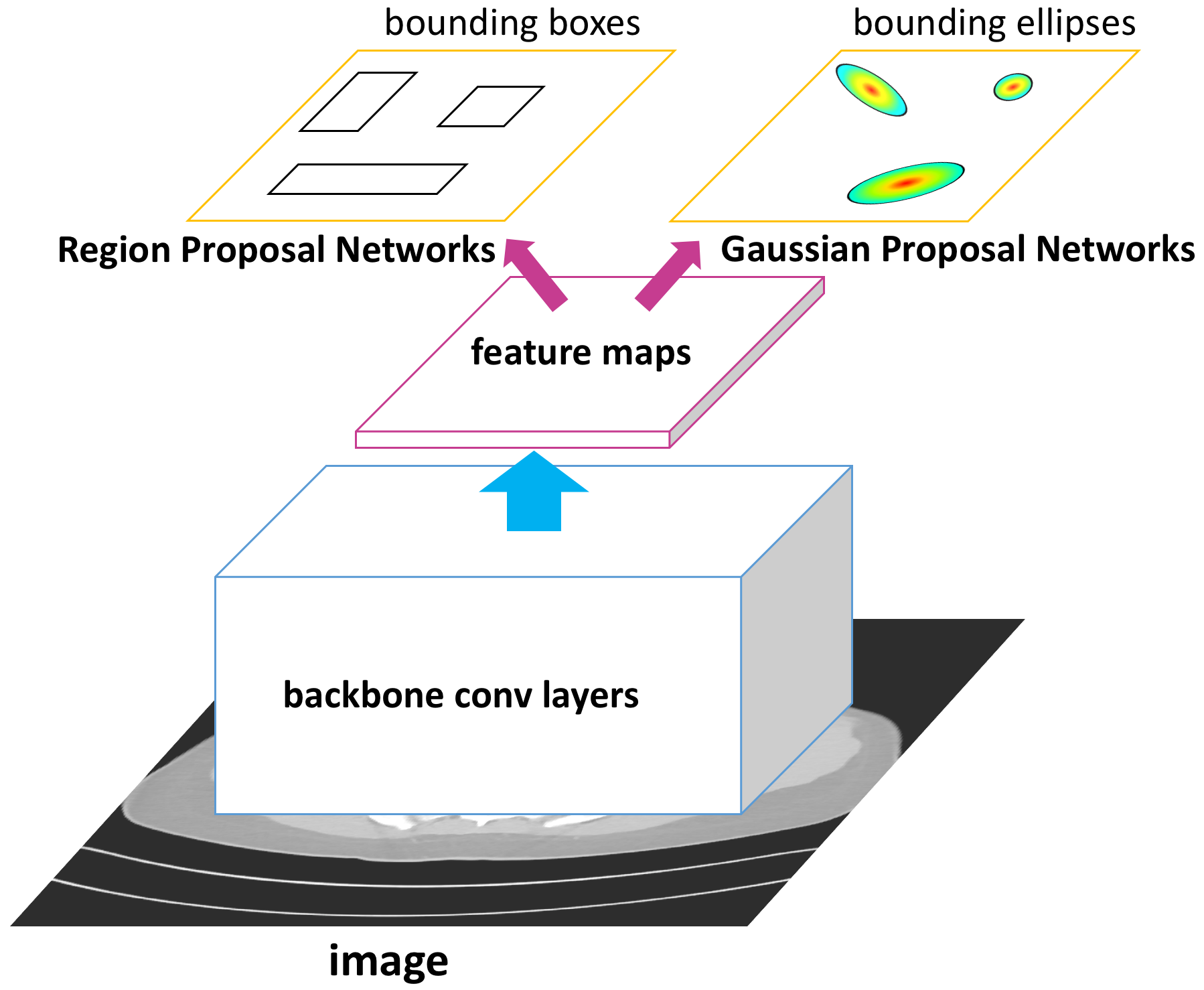}
\end{center}
   \caption{Comparison between Region Proposal Networks and Gaussian Proposal
   Networks. Instead of proposing bounding boxes, Gaussian Proposal Networks
   propose bounding ellipses as 2D Gaussian distributions on the image plane
   and use a single KL divergence loss for object localization.}
\label{fig1}
\end{figure}

\begin{figure}
    \centering
    \begin{subfigure}[b]{0.3\linewidth}
        \includegraphics[width=\linewidth]{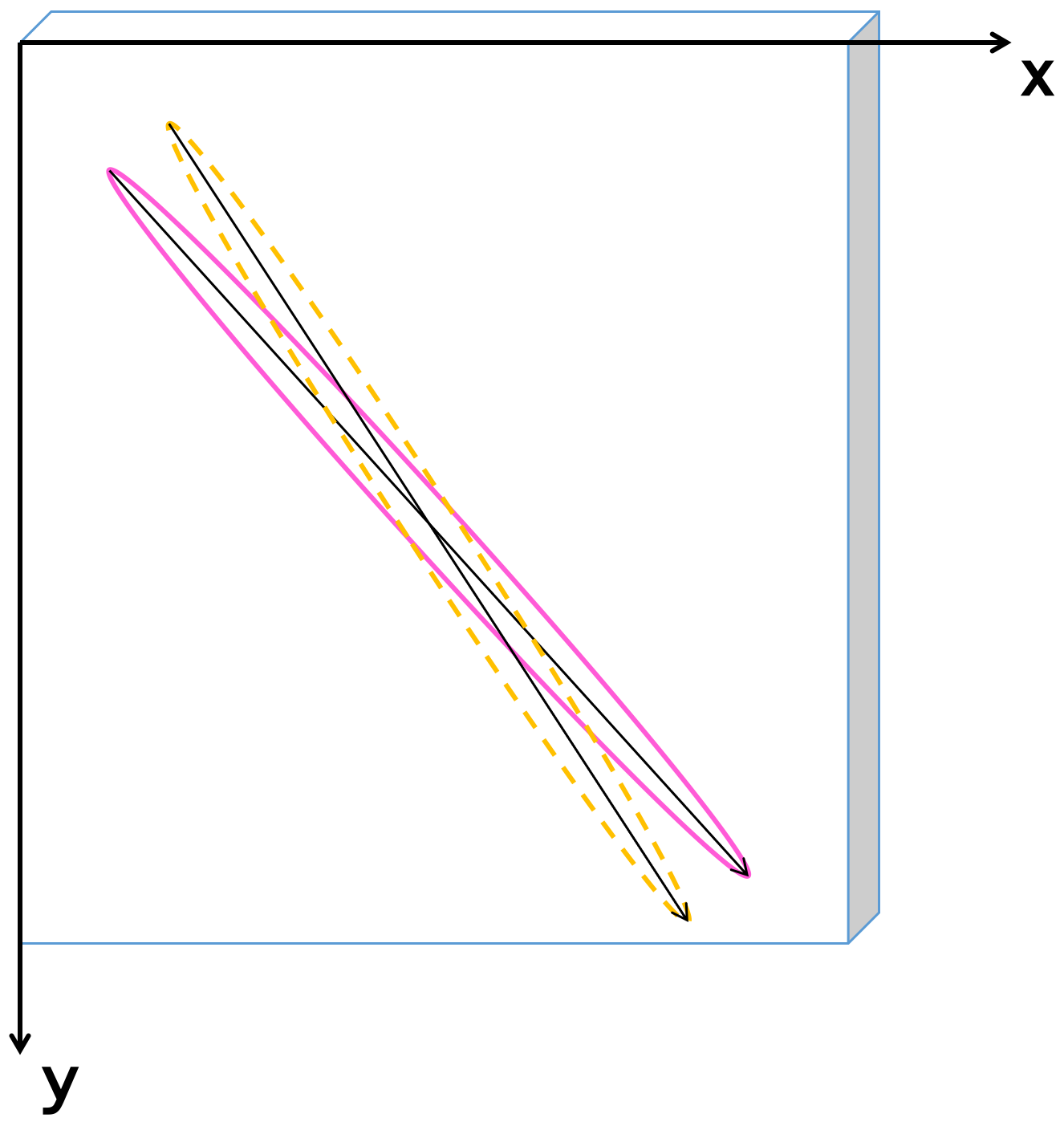}
        \caption{}
        \label{fig2:a}
    \end{subfigure}
    ~ %add desired spacing between images, e. g. ~, \quad, \qquad, \hfill etc. 
    \begin{subfigure}[b]{0.3\linewidth}
        \includegraphics[width=\linewidth]{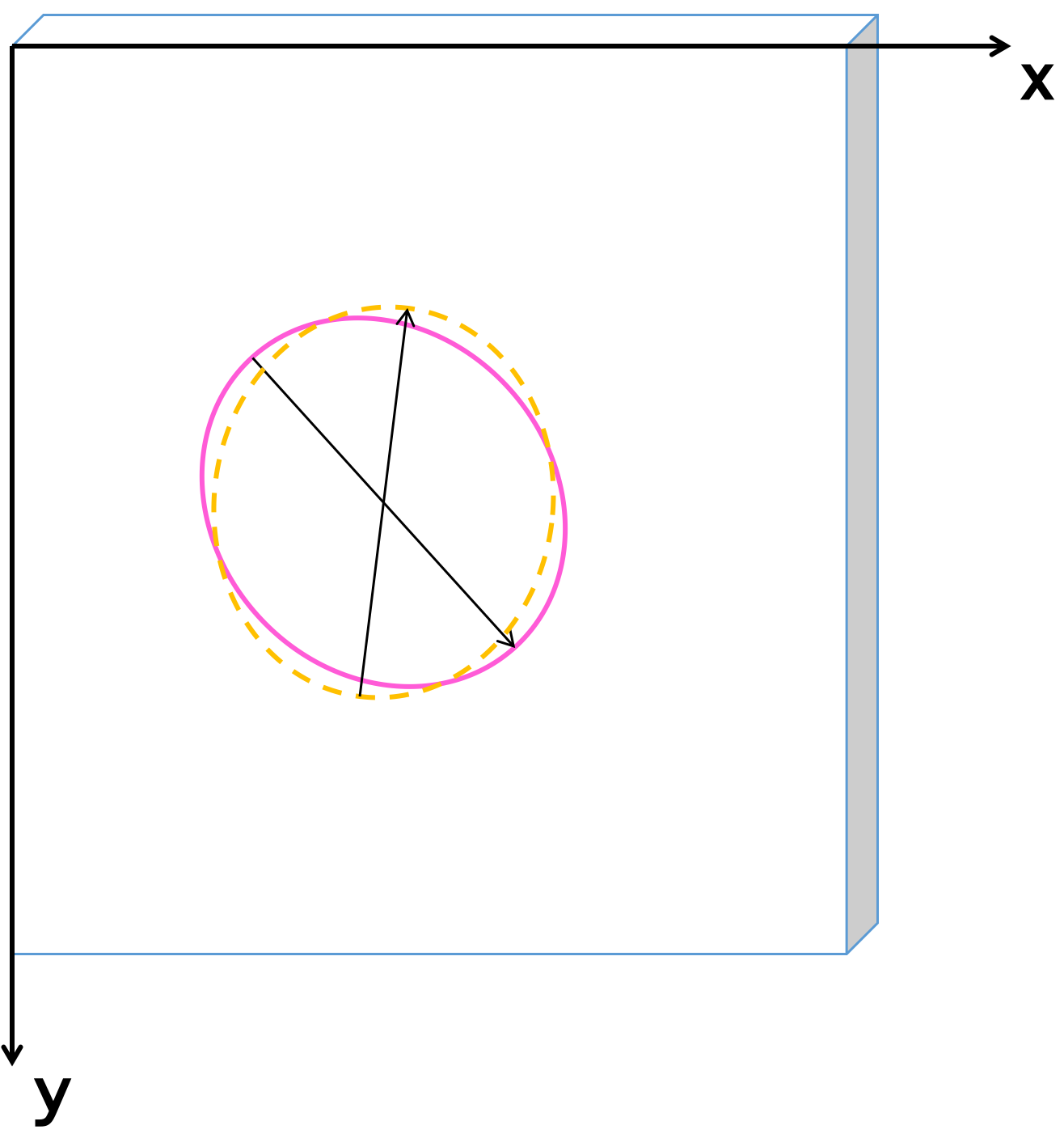}
        \caption{}
        \label{fig2:b}
    \end{subfigure}
    \caption{Two illustrative examples of how the rotation angle may affect the
    overlap between two ellipses differently. The ellipse of solid line is the
    ground truth and the ellipse of dash line is proposed by the detection model,
    where both ellipses only differ in the rotation angle.
    (a) the major axis is significantly longer than the minor axis, (b) the major
    axis is slightly longer than the minor axis.}
    \label{fig2}
\end{figure}

We argue that directly regressing the rotation angle is not optimal for bounding
ellipse localization. The rotation angle and the aspect ratio, i.e. the ratio of
the major axis to the minor axis, jointly affect the overlap between two ellipses.
Figure~\ref{fig2} shows two extreme examples of a proposed bounding ellipse and
its ground truth bounding ellipse where they only differ in the rotation angle.
In Figure~\ref{fig2:a}, where the aspect ratio is significantly larger than 1,
a small shift in the rotation angle results in a dramatic change of overlap between
the two ellipses. However in Figure~\ref{fig2:b}, where the major axis is almost
equal length to the minor axis, the shift of the rotation angle merely affects
the overlap between the two ellipses. Therefore, it may not be optimal or
sometimes even unnecessary to directly regress the rotation angle, when the
essential goal is to optimize the overlap between the proposed ellipse and the
ground truth ellipse.

In this work, we present Gaussian Proposal Networks (GPNs) that learn to propose 
bounding ellipses as 2D Gaussian distributions on the image plane. Figure~\ref{fig1}
shows an illustrative comparison between GPN and RPN. Unlike most of the 
extensions to the RPN framework that introduce an additional term to directly
regress the rotation angle for object localization, GPN minimizes the
Kullback-Leibler (KL) divergence between the proposed Gaussian distribution and
the ground truth Gaussian distribution as one single loss for localization. KL
divergence directly measures the overlap of one distribution with respect to a
reference distribution. When the two distributions are Gaussian, KL divergence
has analytical form and is differentiable. Therefore, GPN can be readily
implemented with standard automatic differentiation packages~\cite{paszke2017automatic}
and trained with back-propagation algorithm. We also show that when the rotation
angle is 0, the KL divergence loss approximately incarnates the regression loss
used in the RPN framework for bounding box localization. We experiment the efficacy
of GPN for detecting lesion bounding ellipses on the DeepLesion~\cite{yan2018deeplesion}
dataset. GPN outperforms RPN by a significant margin in terms
of free-response receiver operating characteristic across different experimental
settings. Error analysis shows that GPN achieves significant lower localization
error compared to RPN. Further analysis on the distribution of predicted rotation
angles from GPN supports our conjecture that it may not be necessary to regress the
rotation angle when the ground truth bounding ellipse has similar lengths of major
and minor axises.

\section{Related work}

%-------------------------------------------------------------------------
\subsection{Region proposal networks (RPNs)}

The design of GPN generally follows the principles of RPN~\cite{ren2015faster}.
RPN has a fully convolutional backbone network that processes the input image and generates
a feature map grid. The feature vector of each position on the feature map is further processed
with a $3\times3$ convolutional layer and then associated with potentially multiple anchors
of varies scales and aspect ratios. The ground truth region of interests (RoIs)
are then assigned to anchors that meet certain overlapping criteria, e.g. intersection
over union (IoU) greater than 0.7. Finally, the $3\times3$ convolutional layer is followed
by two separate $1\times1$ convolutional layers, one is predicting the objectness
scores of the RoIs and the other is predicting the bounding box offsets of the
RoIs with respect to the anchors. RPN is jointly trained with one classification
loss and multiple smoothed L1 regression losses.

Compared to RPN, GPN proposes bounding ellipses as 2D Gaussian distributions and
minimizes a single KL divergence loss between two Gaussian distributions for
localization. GPN could be applied with the same extensions that apply to RPN, 
such as training with multi-scale
feature maps~\cite{lin2017feature, liu2016ssd, fu2017dssd}, online hard negative
mining~\cite{shrivastava2016training} and focal loss~\cite{lin2018focal}. However,
in the Faster R-CNN two-stage detector, a second R-CNN classifier~\cite{ren2015faster}
is appended to RPN through a RoI pooling layer~\cite{girshick2015fast} and classifies
each RoI into specific object categories or background. Performing RoI pooling
with bounding ellipses is nontrivial and is beyond the scope of this paper.
Thus, GPN currently only applies to one-stage detectors like SSD~\cite{liu2016ssd},
where object bounds and categories are jointly predicted through one single
network.

\subsection{Bounding ellipse and KL divergence}
Bounding ellipse annotation has been widely used in human faces detection~\cite{jain2010fddb}.
To generate bounding ellipses, most of the detection methods directly map detected bounding boxes into
ellipses~\cite{li2015convolutional, yang2016multi, mathias2014face, wang2016use, zhang2017s}.
Shi et al.~\cite{shi2018real} discretize the rotation angle and classify it into different
categories. Hu et al.~\cite{hu2017finding} train a separated ellipse
regressor to regress the cotangent of the rotation angle using features extracted
offline from deep networks. Opitz et al.~\cite{opitz2016grid} also point out the problem
that different parameters of bounding ellipses, i.e. the center coordinates,
the major and minor axises and the rotation angle, impact ellipses overlap differently,
thus is in spirit similar to our argument. However, in order to maximize the overlap,
Opitz et al.~\cite{opitz2016grid} rasterize both the proposed ellipse and the ground truth
ellipse and numerically compute the gradient of each ellipse parameter by
counting the change of rasterized overlap in pixels. In comparison, GPN uses KL
divergence to analytically optimize the overlap between two Gaussian distributions
as a surrogate to optimizing ellipses overlap. Najibi et al.~\cite{najibi2018towards}
also use Gaussian distributions to generate the saliency maps of objects based
on their bounding boxes. But they do not consider the rotation angle and do not
use KL divergence loss for optimization.

KL divergence has been used to measure the ellipticity and circularity of a given
set of points~\cite{misztal2016ellipticity}. To the best of our knowledge, the
most similar work to our approach is the recent Softer-NMS~\cite{he2018softer}
that also uses KL divergence for bounding box localization. However, the motivation
of Softer-NMS is to model the location uncertainty of each corner of proposed
bounding boxes through 1D Gaussian distribution. The motivation of GPN is to
propose bounding ellipses as 2D Gaussian distributions on the image plain.
Therefore, it is not directly comparable to Softer-NMS. We discuss the differences
and connections between Softer-NMS and GPN with more details in the Supplementary.

%------------------------------------------------------------------------
\section{Gaussian Proposal Networks (GPNs)}
This section describes the mathematical and implementation details of GPN. We
first formulate the representation of ellipses as 2D Gaussian distributions in
Section~\ref{sec1.1}. We then derive the KL divergence between 2D Gaussian
distributions using this representation in Section~\ref{sec1.2}. Next we draw
connections between the KL divergence loss and the regression loss used in the RPN
framework in Section~\ref{sec1.3}. Finally, we provide some implementation details
in Section~\ref{sec1.4}.

%-------------------------------------------------------------------------
\subsection{Ellipses as 2D Gaussian distributions}\label{sec1.1}
The equation of an ellipse in a 2D coordinate system without rotation is given by
\begin{eqnarray}\label{eq1}
\frac{(x-\mu_x)^2}{\sigma^2_x} + \frac{(y-\mu_y)^2}{\sigma^2_y} = 1,
\end{eqnarray}
where we denote $\mu_x, \mu_y$ as the center coordinates of the ellipse, and $\sigma_x, \sigma_y$
as the lengths of semi-axises along $x$ and $y$ axises.

The probability density
function of a 2D Gaussian distribution is given by
\begin{eqnarray}
f(\mathbf{x}|\boldsymbol{\mu}, \mathbf{\Sigma}) = \frac{ \exp (-\frac{1}{2} (\mathbf{x} - \boldsymbol{\mu})^\intercal \mathbf{\Sigma}^{-1} (\mathbf{x} - \boldsymbol{\mu}) )  }{ 2\pi {|\mathbf{\Sigma}|}^{\frac{1}{2}} },  
\end{eqnarray}
where $\mathbf{x}$ is the vector representation of coordinates $(x, y)$, 
$\boldsymbol{\mu}$ is the mean, $\mathbf{\Sigma}$ is the covariance matrix and
$|\mathbf{\Sigma}|$ is the determinant of the covariance matrix. If we assume
the off-diagonal term in $\mathbf{\Sigma}$ is 0 and parameterize $\boldsymbol{\mu}, \mathbf{\Sigma}$ as
\begin{eqnarray} 
\boldsymbol{\mu} = 
\begin{bmatrix}
\mu_x \\
\mu_y
\end{bmatrix},
\mathbf{\Sigma} = 
\begin{bmatrix}
\sigma^2_x & 0 \\
0          & \sigma^2_y
\end{bmatrix},
\end{eqnarray}
then the ellipse equation in Equation~\ref{eq1} corresponds to the density contour
of the 2D Gaussian distribution when
\begin{eqnarray}
(\mathbf{x} - \boldsymbol{\mu})^\intercal \mathbf{\Sigma}^{-1} (\mathbf{x} - \boldsymbol{\mu}) = 1.
\end{eqnarray}

\begin{figure}[t]
\begin{center}
   \includegraphics[width=0.5\linewidth]{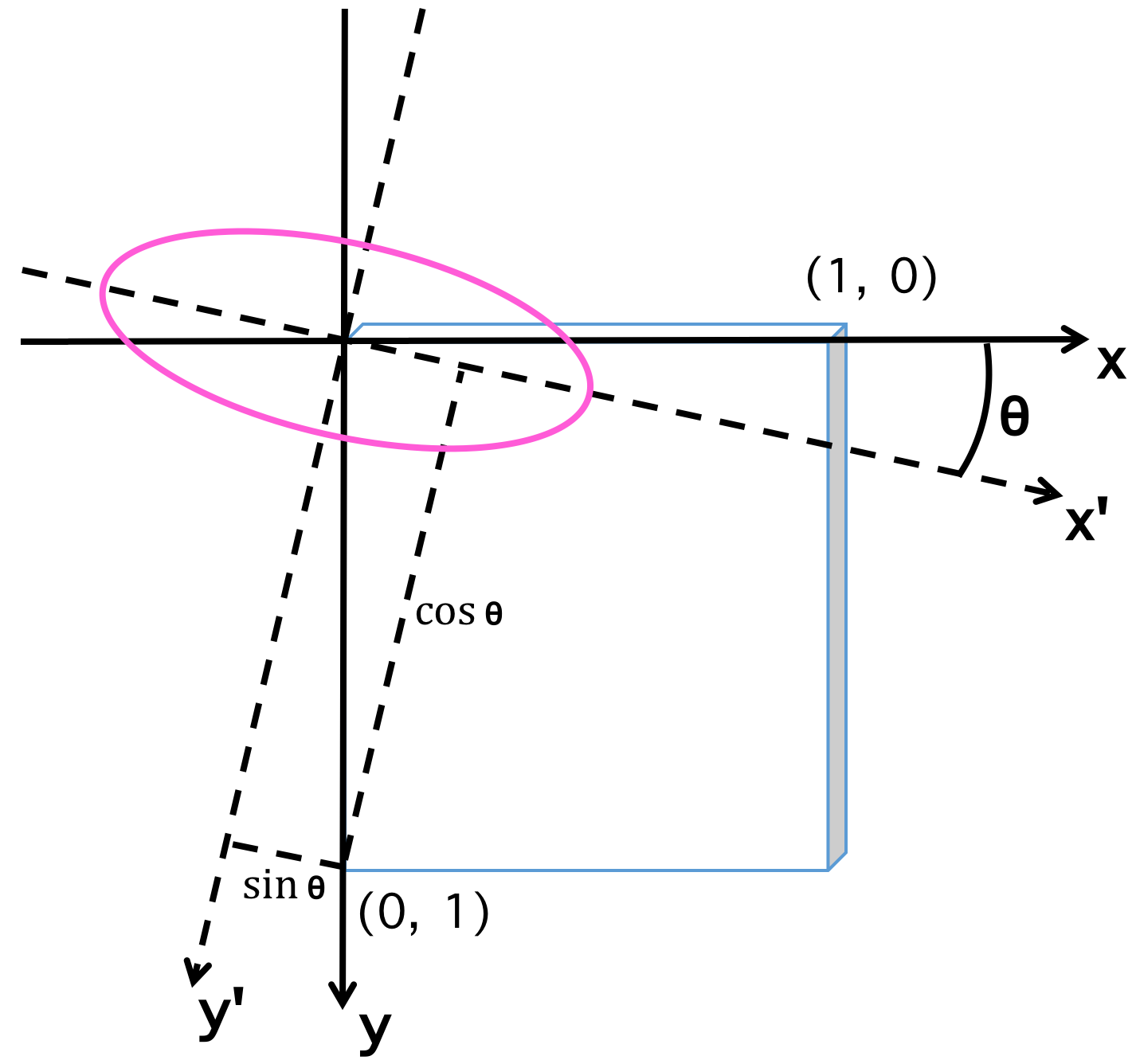}
\end{center}
   \caption{Correspondence between the original coordinate system $(x, y)$, and
   the rotated coordinate system $(x^\prime, y^\prime)$ with a rotation angle
   $\theta$. The major axis of the ellipse is aligned with the $x^\prime$ axis
   in the rotated system. The coordinates of $(1, 0)$ and $(0, 1)$ in the
   $(x, y)$ system, are $(\cos \theta, -\sin \theta)$ and $(\sin \theta, \cos \theta)$
   in the $(x^\prime, y^\prime)$ system.}
\label{fig3}
\end{figure}

When the major axis of the ellipse is rotated of an angle $\theta$ with respect to
the $x$ axis, we use a rotation matrix $R(\theta)$ to map the coordinates in the
original $(x, y)$ system into a new $(x^\prime, y^\prime)$ system, where the major
axis of the ellipse is aligned with the $x^\prime$ axis in the new system as shown
in Figure~\ref{fig3}

\begin{eqnarray}
\begin{bmatrix}
x^\prime \\
y^\prime
\end{bmatrix} = R(\theta)
\begin{bmatrix}
x \\
y
\end{bmatrix} ,
R(\theta) = 
\begin{bmatrix}
\cos \theta   & \sin \theta \\
-\sin \theta  & \cos \theta
\end{bmatrix}.
\end{eqnarray}
For example, the coordinates of $(1, 0)$ and $(0, 1)$ in the $(x, y)$ system,
are $(\cos \theta, -\sin \theta)$ and $(\sin \theta, \cos \theta)$ in the
$(x^\prime, y^\prime)$ system. If we use $\sigma_l, \sigma_s$ to denote the
lengths of the semi-major (long) and semi-minor (short) axises of the ellipse in the
$(x^\prime, y^\prime)$ system and assume it is centered at $(\mu_x^\prime, \mu_y^\prime)$,
then the ellipse equation in the $(x^\prime, y^\prime)$ system is given by
\begin{eqnarray}
(\mathbf{x}^\prime - \boldsymbol{\mu}^\prime)^\intercal (\mathbf{\Sigma}^\prime)^{-1} (\mathbf{x}^\prime - \boldsymbol{\mu}^\prime) = 1,
(\mathbf{\Sigma}^\prime)^{-1} = 
\begin{bmatrix}
\frac{1}{\sigma^2_l} & 0 \\
0                    & \frac{1}{\sigma^2_s}
\end{bmatrix},
\end{eqnarray}
and again, $\mathbf{x}^\prime$ and $\boldsymbol{\mu}^\prime$ are the vector representations
of $(x^\prime, y^\prime)$ and $(\mu_x^\prime, \mu_y^\prime)$.

With the rotation matrix $R(\theta)$, the ellipse equation in the $(x, y)$ system 
can be derived as
\begin{equation}
\begin{aligned}
& [R(\theta) (\mathbf{x} - \boldsymbol{\mu})]^\intercal (\mathbf{\Sigma}^\prime)^{-1} [R(\theta) (\mathbf{x} - \boldsymbol{\mu})] = 1, \\
& (\mathbf{x} - \boldsymbol{\mu})^\intercal [R^\intercal(\theta) (\mathbf{\Sigma}^\prime)^{-1} R(\theta)] (\mathbf{x} - \boldsymbol{\mu}) = 1,
\end{aligned}
\end{equation}
where $\boldsymbol{\mu}$ is the center coordinates $(\mu_x, \mu_y)$ of the ellipse
in the $(x, y)$ system.

Finally, we can use a 2D Gaussian distribution in the $(x, y)$ system parameterized by
\begin{eqnarray}\label{eq2}
\boldsymbol{\mu} = 
\begin{bmatrix}
\mu_x \\
\mu_y
\end{bmatrix},
\mathbf{\Sigma}^{-1} = R^\intercal(\theta)
\begin{bmatrix}
\frac{1}{\sigma^2_l} & 0 \\
0                    & \frac{1}{\sigma^2_s}
\end{bmatrix} R(\theta)
\end{eqnarray}
to represent the ellipse centered at $(\mu_x, \mu_y)$, with semi-major and semi-minor
axises of lengths $(\sigma_l, \sigma_s)$, and a rotation angle of $\theta$ between
its major axis and the $x$ axis. Note that $\theta \in [-\frac{\pi}{2}, \frac{\pi}{2}]$.

The goal of GPN is to propose bounding ellipses such that their parameters,
$(\mu_x, \mu_y, \sigma_l, \sigma_s, \theta)$, match the ground truth ellipses
through the criteria of KL divergence.

\subsection{KL divergence of 2D Gaussian distributions}\label{sec1.2}
The KL divergence between a proposed 2D Gaussian distribution $\mathcal{N}_p$
and a target 2D Gaussian distribution $\mathcal{N}_t$ is given by~\cite{duchi2007derivations}
\begin{equation}\label{eq3}
\begin{aligned}
  & D_{\mathrm{KL}}(\mathcal{N}_t||\mathcal{N}_p) = \frac{1}{2} \bigg[ \mathrm{tr}(\mathbf{\Sigma}^{-1}_p \mathbf{\Sigma}_t)+(\boldsymbol{\mu}_p - \boldsymbol{\mu}_t)^\intercal \mathbf{\Sigma}^{-1}_p (\boldsymbol{\mu}_p - \boldsymbol{\mu}_t) + \mathrm{ln} \frac{|\mathbf{\Sigma}_p|}{|\mathbf{\Sigma}_t|} - 2 \bigg], 
\end{aligned}
\end{equation}
where $\mathrm{tr}(\mathbf{X})$ is the trace of matrix $\mathbf{X}$.

Assuming $\mathcal{N}_p$ and $\mathcal{N}_t$ are parameterized by
$(\mu_{x_p}, \mu_{y_p}, \sigma_{l_p}, \sigma_{s_p}, \theta_p)$ and 
$(\mu_{x_t}, \mu_{y_t}, \sigma_{l_t}, \sigma_{s_t}, \theta_t)$ following
Equation~\ref{eq2}, we can derive each term in Equation~\ref{eq3} as
\begin{equation}\label{eq4}
\begin{aligned}
  & \mathrm{tr}(\mathbf{\Sigma}^{-1}_p \mathbf{\Sigma}_t) = \cos^2\!\Delta\theta \ \frac{\sigma^2_{l_t}}{\sigma^2_{l_p}} + \cos^2\!\Delta\theta \ \frac{\sigma^2_{s_t}}{\sigma^2_{s_p}} + \sin^2\!\Delta\theta \ \frac{\sigma^2_{l_t}}{\sigma^2_{s_p}} + \sin^2\!\Delta\theta \ \frac{\sigma^2_{s_t}}{\sigma^2_{l_p}},
\end{aligned}
\end{equation}

\begin{equation}\label{eq5}
\begin{aligned}
  & (\boldsymbol{\mu}_p - \boldsymbol{\mu}_t)^\intercal \mathbf{\Sigma}^{-1}_p (\boldsymbol{\mu}_p - \boldsymbol{\mu}_t) = \frac{(\cos\theta_p \Delta x + \sin\theta_p \Delta y)^2}{\sigma^2_{l_p}} + \frac{(\cos\theta_p \Delta y - \sin\theta_p \Delta x)^2}{\sigma^2_{s_p}},
\end{aligned}
\end{equation}

\begin{eqnarray}\label{eq6}
\mathrm{ln} \frac{|\mathbf{\Sigma}_p|}{|\mathbf{\Sigma}_t|} = \mathrm{ln} \frac{\sigma^2_{l_p}}{\sigma^2_{l_t}} + \mathrm{ln} \frac{\sigma^2_{s_p}}{\sigma^2_{s_t}},
\end{eqnarray}
where we define
\begin{eqnarray}
\Delta\theta = \theta_p -\theta_t, \Delta x = \mu_{x_p} - \mu_{x_t}, \Delta y = \mu_{y_p} - \mu_{y_t}.
\end{eqnarray}
The exact details of deriving each term in Equation~\ref{eq4} through Equation~\ref{eq6}
are provided in the Supplementary.

\subsection{Connection to the RPN regression loss}\label{sec1.3}
The general form of KL divergence derived in the previous section looks rather
complex. However, if we omit the rotation angle, i.e. assuming $\theta_p$ and
$\theta_t$ are always 0, then $\sigma_l$ and $\sigma_s$ are actually the half
width and the half height of the bounding box that tightly surrounds the bounding
ellipse. And we obtain a much simpler form of KL divergence as
\begin{equation}\label{eq7}
\begin{aligned}
  & D_{\mathrm{KL}}(\mathcal{N}_t||\mathcal{N}_p) = \frac{1}{2} \bigg[ \frac{w^2_t}{w^2_p} + \frac{h^2_t}{h^2_p} + \frac{\Delta^2 x}{w^2_p} + \frac{\Delta^2 y}{h^2_p} + \mathrm{ln} \frac{w^2_p}{w^2_t} + \mathrm{ln} \frac{h^2_p}{h^2_t} - 2 \bigg].
\end{aligned}
\end{equation}
where we have replaced $(\sigma_{l_p}, \sigma_{l_t})$ with $(w_p, w_t)$, and
$(\sigma_{s_p}, \sigma_{s_t})$ with $(h_p, h_t)$ for easy comparison with the RPN
regression loss.

For bounding box regression, RPN outputs four terms for each proposed bounding box~\cite{ren2015faster}
\begin{equation}\label{eq8}
\begin{aligned}
& t^p_x = (x_p - x_a) / w_a , \ \ t^p_y = (y_p - y_a) / h_a, \\
& t^p_w = \mathrm{ln} (w_p/w_a) , \ \ t^p_h = \mathrm{ln} (h_p/h_a),
\end{aligned}
\end{equation}
to match the four targets from the ground truth bounding box
\begin{equation}\label{eq9}
\begin{aligned}
& t^t_x = (x_t - x_a) / w_a , \ \ t^t_y = (y_t - y_a) / h_a, \\
& t^t_w = \mathrm{ln} (w_t/w_a) , \ \ t^t_h = \mathrm{ln} (h_t/h_a),
\end{aligned}
\end{equation}
where $(x_a, y_a, w_a, h_a)$ are the center coordinates, width and height of the
matching anchor.

RPN uses smoothed L1 loss for regression. When the loss is small, smoothed L1
loss becomes squared loss. Therefore, for center coordinates regression, the
squared loss are
\begin{eqnarray}
 (t^p_x - t^t_x)^2 = \frac{\Delta^2 x}{w^2_a}, \ \ (t^p_y - t^t_y)^2 = \frac{\Delta^2 y}{h^2_a},
\end{eqnarray}
which are very similar to the middle two terms in Equation~\ref{eq7}, except that
the RPN regression loss normalizes $(\Delta x, \Delta y)$ by the anchor width and height
$(w_a, h_a)$, where KL divergence normalizes them by the predicted width and height 
$(w_p, h_p)$.

When the loss is large, the L1 losses for width and height regression are
\begin{eqnarray}
|t^p_w - t^t_w| = \left|\mathrm{ln} \frac{w_p}{w_t}\right|, \ \ |t^p_h - t^t_h| = \left|\mathrm{ln} \frac{h_p}{h_t}\right|,
\end{eqnarray}
which are equivalent to the last two terms (ignoring the constant $-2$) in Equation~\ref{eq7}
when $w_p \ge w_t, h_p \ge h_t$. However, since KL divergence is asymmetric,
when $w_p < w_t, h_p < h_t$, KL divergence penalizes differently through the first
two terms in Equation~\ref{eq7}.

The comparison between the KL divergence loss and the smoothed L1 regression loss in
RPN, not only provides another perspective towards the efficacy of RPN for object
localization, but also suggests an alternative loss for bounding box localization,
i.e. Equation~\ref{eq7}. However, the efficacy of the KL divergence loss for general
bounding box localization needs comprehensive analysis on other benchmark datasets,
e.g. PASCAL VOC~\cite{everingham2010pascal} and MS COCO~\cite{lin2014microsoft},
and is beyond the scope of this paper.

\subsection{Implementation details}\label{sec1.4}
To implement the KL divergence loss in practice, we follow the design of anchors
from RPN and make the four targets the same as Equation~\ref{eq9} based on the
matching anchor, plus the tangent of the rotation angle following~\cite{hu2017finding}.
We let GPN output four terms the same as Equation~\ref{eq8}, plus an additional
term for the tangent of the rotation angle. Finally, we plug both the network
outputs and the targets into Equation~\ref{eq3} through Equation~\ref{eq6} to
compute the KL divergence loss.

The KL divergence loss can be directly added with the classification loss without
any balancing factors used in RPN
\begin{equation}
\begin{aligned}
L_\text{total} = L_{\text{cls}} + L_{\text{KLD}}. 
\end{aligned}
\end{equation}
The only important caveat we found to make the KL divergence loss well bounded is
to initialize the weights of the $1\times1$ convolutional layer for bounding ellipse
localization within a small range. Specifically, we use a Gaussian distribution
with 0 mean and standardization of 0.001 to initialize the weights.

GPN was implemented with PyTorch-0.3.1~\cite{paszke2017automatic}.

%------------------------------------------------------------------------
\section{Experiments}
We present comprehensive evaluation of GPN for detecting lesion bounding ellipses on
the DeepLesion dataset~\cite{yan2018deeplesion}. We first introduce some details
about the DeepLesion dataset in Section~\ref{sec2.1}, and experiments setup in
Section~\ref{sec2.2}. Next, we show GPN significantly outperforms RPN for bounding
ellipse detection across different settings in Section~\ref{sec2.3}. Finally, we
present a comprehensive error analysis in Section~\ref{sec2.4}.

%-------------------------------------------------------------------------
\subsection{DeepLesion dataset}\label{sec2.1}

\begin{figure}[t]
\begin{center}
   \includegraphics[width=0.5\linewidth]{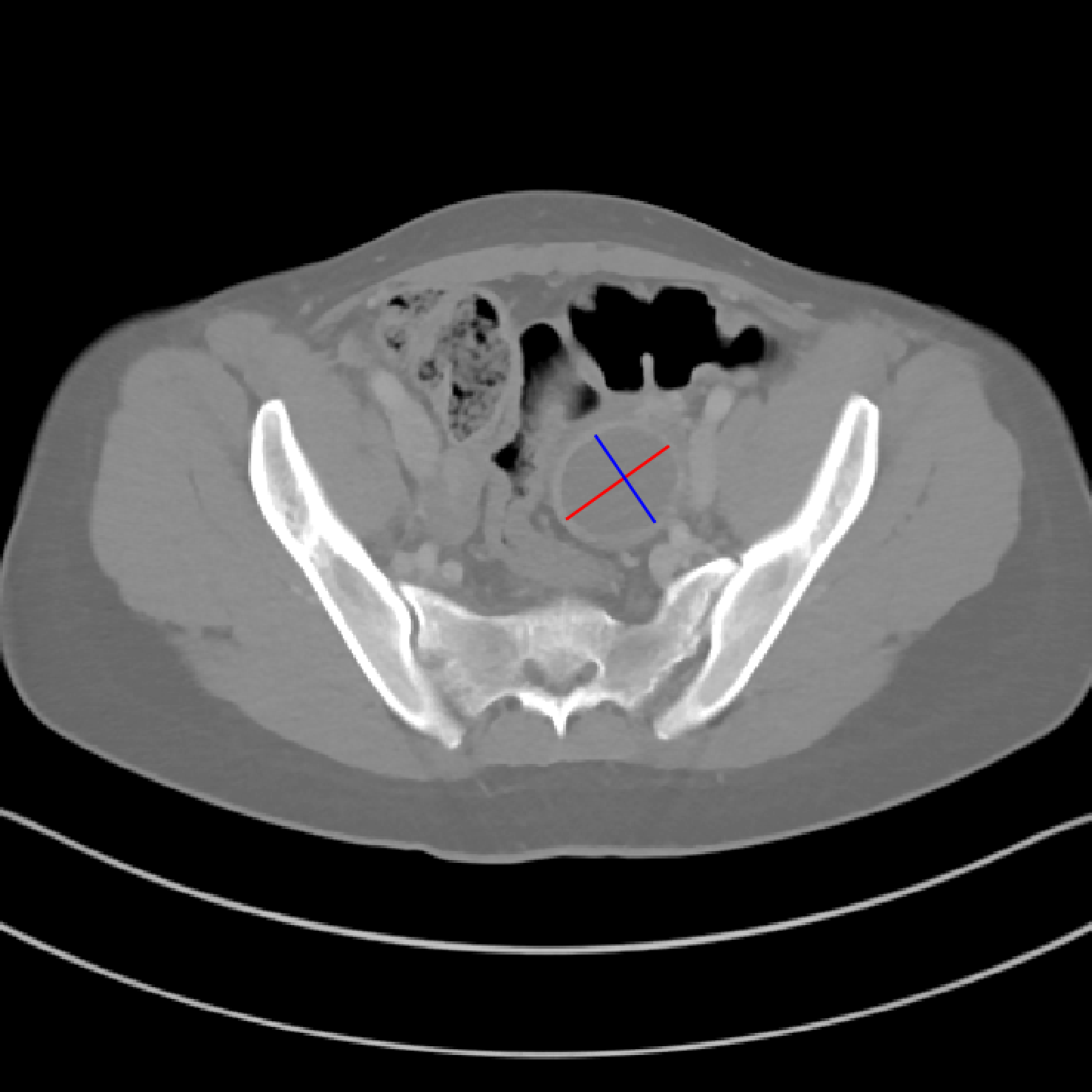}
\end{center}
   \caption{One example of slice image and its RECIST diameters annotation from the
   DeepLesion dataset~\cite{yan2018deeplesion}. The red line is the major axis
   that measures the longest diameter of the annotated lesion and the blue line
   is the minor axis that measures the longest diameter perpendicular to the major axis.}
\label{fig4}
\end{figure}

\begin{figure}
    \centering
    \begin{subfigure}[b]{0.48\linewidth}
        \includegraphics[width=\linewidth]{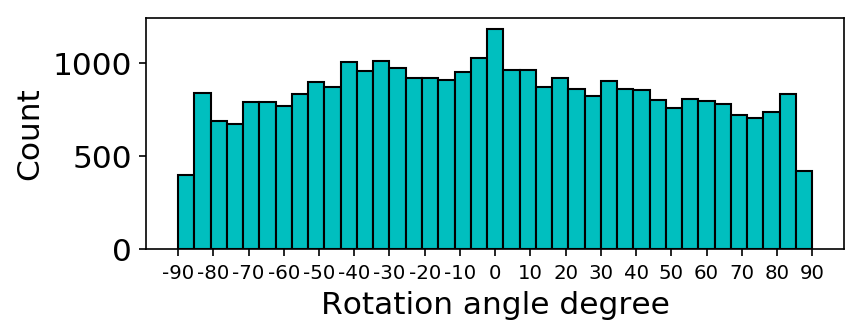}
        \caption{}
        \label{fig5:a}
    \end{subfigure}
    ~
    \begin{subfigure}[b]{0.48\linewidth}
        \includegraphics[width=\linewidth]{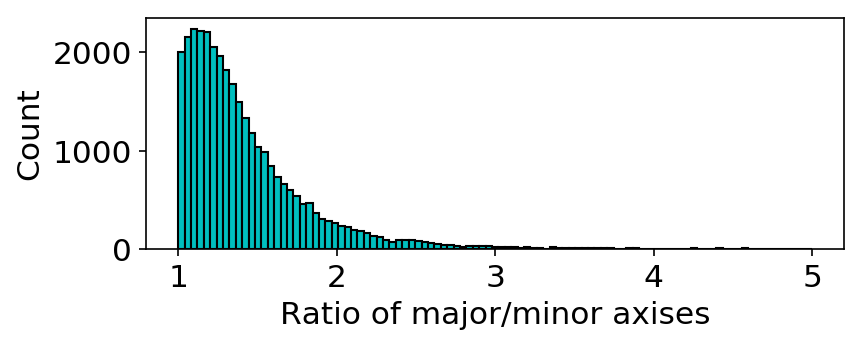}
        \caption{}
        \label{fig5:b}
    \end{subfigure}
    \caption{(a) the distribution of the rotation angles between the lesion's major
    axis and the x axis in degree, (b) the distribution of lesions' aspect ratios.}
    \label{fig5}
\end{figure}

DeepLesion is a large-scale medical imaging dataset recently released from
NIH~\cite{yan2018deeplesion}. It contains 32,735 lesions in 32,120 CT slice
images from 4,427 unique patients. More than $99\%$ of the slice images are 512$\times$512
pixels. Each lesion is annotated with two response evaluation criteria in solid
tumors (RECIST) diameters. The first one measures the longest diameter of the
lesion and the second one measures the longest diameter perpendicular to the
first diameter, so they closely represent the major and minor axises of a bounding
ellipse, and we use this notion thereafter. We note this assumption may be
inaccurate when the center of the minor axis is not aligned with the center of
the major axis. However, we found that for more than $90\%$ of the RECIST-diameter
annotations, the center of the minor axis is within the middle $20\%$ range of
the major axis, so we think the assumption approximately holds. Figure~\ref{fig4}
shows an example of slice image and its RECIST-diameter annotation. DeepLesion
has a wide range of rotation angles and aspect ratios, therefore it is particularly
challenge for bounding ellipse detection and localization. Figure~\ref{fig5:a}
shows the distribution of the rotation angles between the lesion's major axis
and the x axis. Figure~\ref{fig5:b} shows the distribution of lesions' aspect
ratios. For more details about the DeepLesion dataset please refer to~\cite{yan2018deeplesion}.

\subsection{Experiments setup}\label{sec2.2}
We follow the practices from~\cite{yan20183d} to convert the raw slice images
with pixel value in Hounsfield Unit (HU) into 512$\times$512 three channel images with pixel
values between 0 and 255. We use the official split from DeepLesion for training ($70\%$), validation ($15\%$),
and test ($15\%$). All the networks are trained with 20 epochs. We compute intersection
over union (IoU) between ellipses by rasterizing ellipses first and then counting the
pixel overlaps. However, this numerical approach is compute intensive, therefore
we only use it for performance evaluation. During training, we use the bounding box
that tightly surrounds the bounding ellipse to compute IoU for anchor assignment
and non-maximum suppression. Complete details about data preprocessing and
model training are provided in the Supplementary.

\subsection{Performances for bounding ellipse detection}\label{sec2.3}
We evaluate the performances of GPN and RPN for bounding ellipse detection using
a single-scale feature map of stride 8 and five anchor scales, i.e. $(16, 24, 32, 48, 96)$,
following~\cite{yan20183d}. Compared to the Faster R-CNN two-stage detector used in~\cite{yan20183d}, both
GPN and RPN are one-stage detectors that suffer from overwhelming number of background
proposals during training~\cite{lin2018focal}. Therefore, we also experiment training
with just one anchor scale of 16 to mitigate this issue. Both settings are equally
applied to GPN and RPN except that GPN uses the
KL divergence loss for localization while RPN uses the default smoothed L1 loss
for localization with the additional term to regress the tangent of the rotation
angle~\cite{hu2017finding}. We use pretrained VGG-16~\cite{simonyan2014very}
as the backbone network following~\cite{yan20183d} and a single anchor aspect
ratio of 1:1.

\begin{figure}
    \centering
    \begin{subfigure}[b]{0.48\linewidth}
        \includegraphics[width=\linewidth]{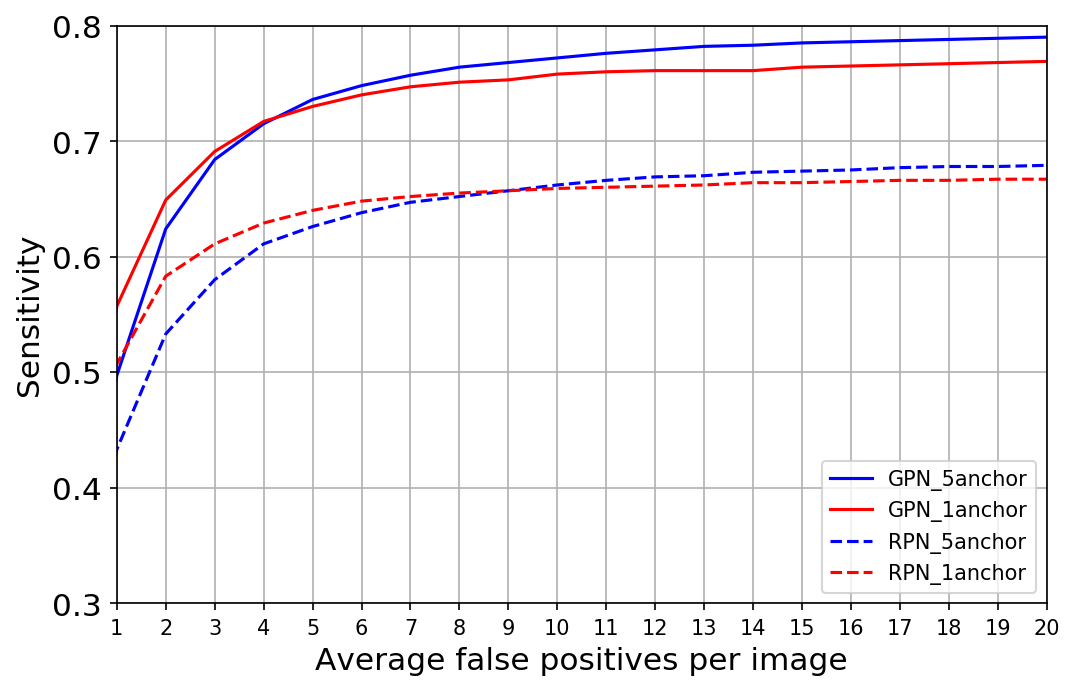}
        \caption{}
        \label{fig6:a}
    \end{subfigure}
    ~
    \begin{subfigure}[b]{0.48\linewidth}
        \includegraphics[width=\linewidth]{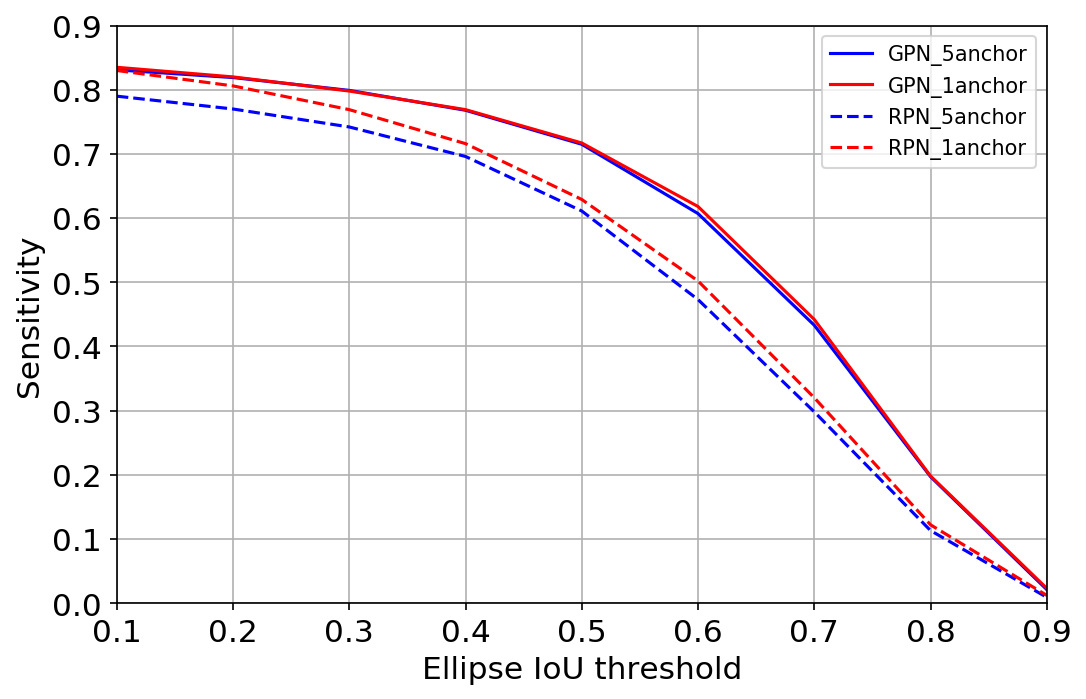}
        \caption{}
        \label{fig6:b}
    \end{subfigure}
    \caption{(a) FROC curves of GPN and RPN on the test set of DeepLesion at 0.5 ellipse
     IoU threshold. (b) Detection sensitivities of GPN and RPN on the test set of DeepLesion
     with different ellipse IoU thresholds at 4 false positives per image.}
    \label{fig6}
\end{figure}

\begin{table}
\begin{center}
\begin{tabular}{|l|c|c|c|c|c|c|}
\hline
FPs per image & 0.5  & 1    & 2    & 4    & 8    & 16 \\
\hline\hline
GPN-5anchor  & 0.36 & 0.50 & 0.62 & 0.72 & 0.76 & 0.79 \\
RPN-5anchor  & 0.31 & 0.43 & 0.53 & 0.61 & 0.65 & 0.68 \\
\hline
GPN-1anchor  & 0.45 & 0.56 & 0.65 & 0.72 & 0.75 & 0.77 \\
RPN-1anchor  & 0.41 & 0.51 & 0.58 & 0.63 & 0.66 & 0.67 \\
\hline
Faster R-CNN  & 0.57 & 0.67 & 0.76 & 0.82 & 0.86 & 0.89 \\
\hline
\end{tabular}
\end{center}
\caption{Detection sensitivities of GPN and RPN on the test set of DeepLesion
at different false positives per image. Ellipse IoU of 0.5 is used as the threshold.
The Faster R-CNN results are from~\cite{yan20183d} where the IoU is computed based on bounding box.}
\label{tab1}
\end{table}

Figure~\ref{fig6:a} and Table~\ref{tab1} show the overall performances of GPN
and RPN on the test set of DeepLesion across different settings measured by the
free-response receiver operating characteristic (FROC). FROC measures the detection
sensitivity with different average false positives per image. We consider a proposed
bounding ellipse is correct if its IoU with the ground truth ellipse is greater
or equal than 0.5. Note that, IoU between ellipses is a very stringent criteria,
especially when the ellipse aspect ratio is significantly larger than one as
illustrated by Figure~\ref{fig2:a}. 

GPN consistently outperforms RPN across both settings by a significant
margin. We also include the previous state-of-the-art results on the DeepLesion
dataset based on Faster R-CNN~\cite{yan20183d} in Table~\ref{tab1}. Yet, the
Faster R-CNN results were trained and evaluated on the bounding box that tightly
surrounds the bounding ellipse, so it is not directly comparable to our results.
One anchor scale training improve the performances of GPN when the average false
positives per image is less or equal than 3.

\subsection{Error analysis}\label{sec2.4}

\begin{figure*}[t]
\begin{center}
   \includegraphics[width=0.98\linewidth]{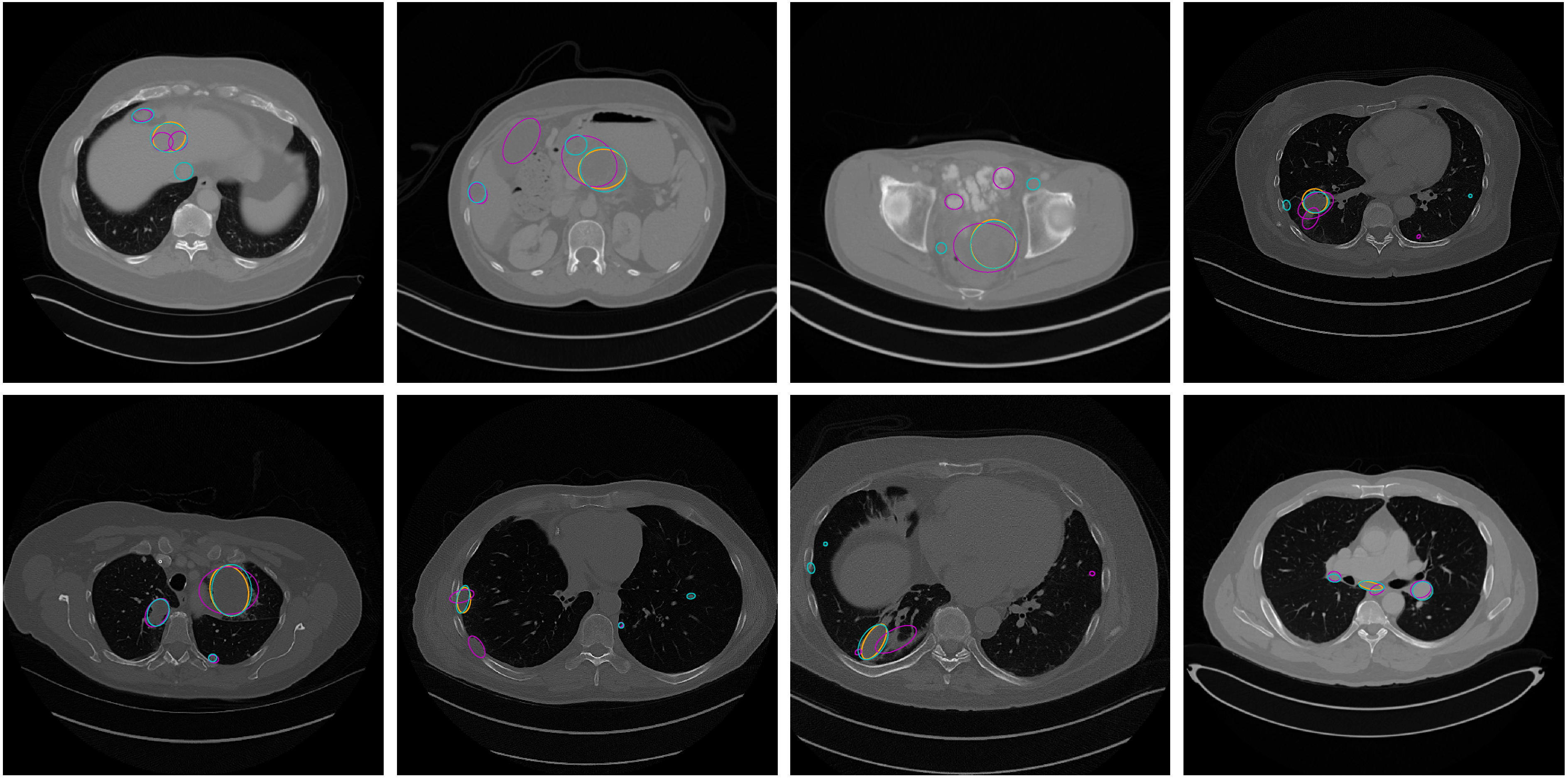}
\end{center}
   \caption{Selected examples of proposed bounding ellipses from GPN-5anchor (cyan) and
   RPN-5anchor (magenta) compared to the ground truth (orange) on the test set of
   DeepLesion. Only the top 3 proposed ellipses with the highest classification
   scores from each model are selected for each image.}
\label{fig7}
\end{figure*}

Figure~\ref{fig7} shows a few examples of proposed bounding ellipses from GPN-5anchor and
RPN-5anchor compared to the ground truth on the test set of DeepLesion. To focus
comparison on ellipse localization, we only select proposed bounding ellipses
that are overlapped with the ground truth by both models. We can see that GPN
detects ellipses of various sizes, rotation angles and aspect ratios
with more accurate overlaps than RPN. We present detailed error analysis in the
next two sections.

\subsubsection{Localization error}
We first investigate the contribution of localization error to detection performance.
Figure~\ref{fig6:b} shows the detection sensitivities of GPN and RPN with different
ellipse IoU thresholds at 4 false positives per image. We can see that, when the 
IoU threshold is small, both GPN and RPN have comparable detection sensitivities
since it is dominated by proposals vs background classification accuracy. As the
IoU threshold increases, the performance of RPN decreases significantly faster
than GPN, suggesting its localization error is significantly higher than GPN.

\subsubsection{Rotation angle error}

\begin{figure}[t]
\begin{center}
   \includegraphics[width=0.8\linewidth]{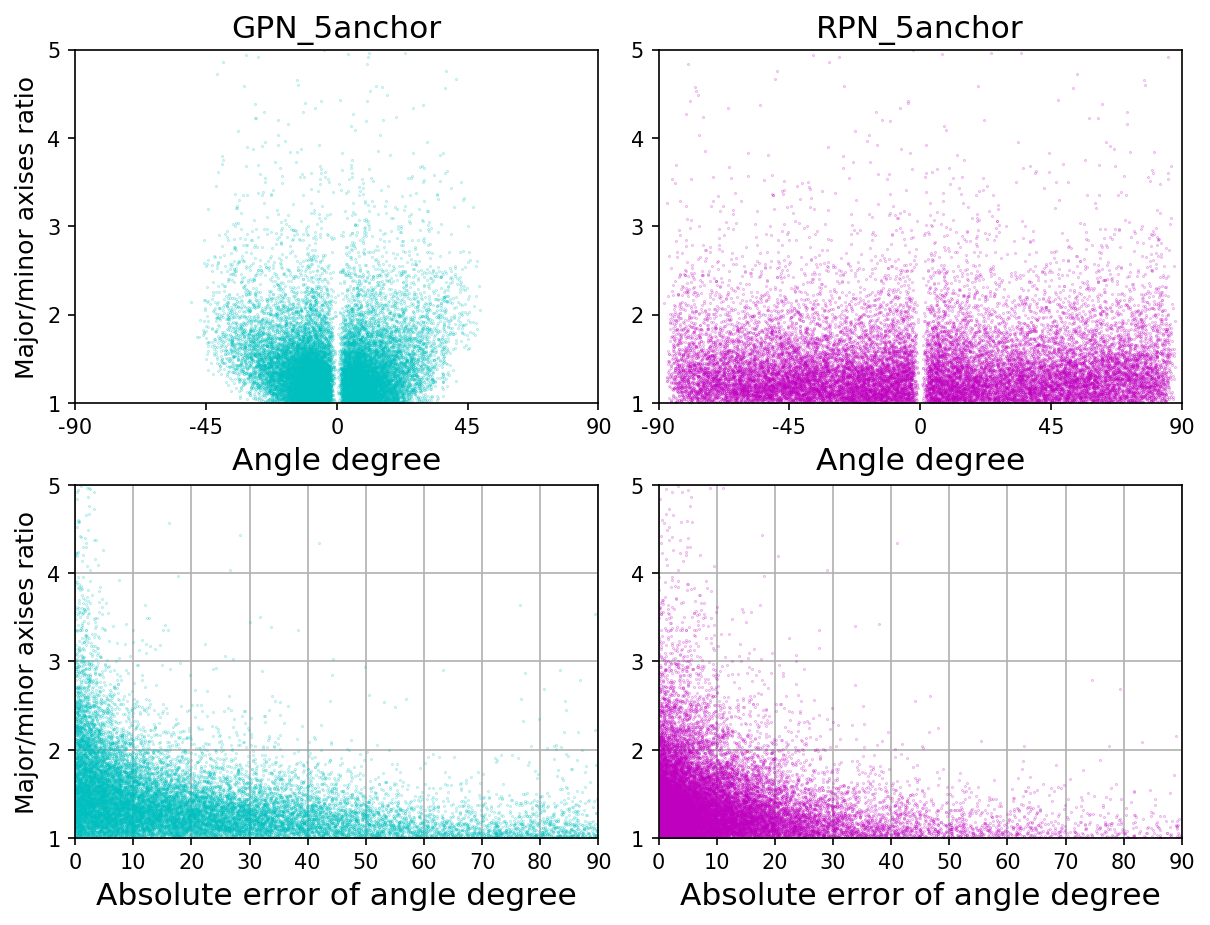}
\end{center}
   \caption{\textbf{The upper panels} are the distributions of predicted angles of
   GPN-5anchor and RPN-5anchor with respect to the ground truth aspect ratio
   on the training set of DeepLesion. \textbf{The lower panels} are the
   absolute degree errors of predicted angles of GPN-5anchor and RPN-5anchor with
   respect to the ground truth aspect ratio on the training set of DeepLesion.}
\label{fig9}
\end{figure}

We also investigate the behaviors of rotation angle prediction of GPN and RPN.
In the Introduction Section, we conjecture that it may be unnecessary to directly regress
the rotation angle for bounding ellipse localization, especially when the ellipse
aspect ratio is close to 1 as illustrated in Figure~\ref{fig2:b}. The
upper panels of Figure~\ref{fig9} show the distributions of predicted angles of
GPN-5anchor and RPN-5anchor with respect to the aspect ratio on
the training set of DeepLesion. We notice when the aspect ratio is
close to 1, GPN mostly predicts the rotation angle around 0 since it merely affects the ellipses overlap. On the
other hand, because RPN directly regresses the rotation angle, the distribution
of its predicted angles is much wider regardless of the aspect ratio,
and similar to the ground truth distribution in Figure~\ref{fig5:a}. We notice
the absolute degrees of predicted angles from GPN rarely exceed $45^{\circ}$.
This is because GPN tends to flip the major and minor axises representation when
the ground truth rotation angle of the major axis is greater than $45^{\circ}$
or less than $-45^{\circ}$, since the flipped representation contributes symmetrically
to the KL divergence loss.

The lower panels of Figure~\ref{fig9} show the absolute degree errors of predicted
angles of GPN-5anchor and RPN-5anchor on the training set of DeepLesion. We use
the predicted longer axis as the actual major axis to account for the 
flipping effect of GPN. We see the angle errors of both GPN and RPN are significantly
lower when the ground truth aspect ratio is significantly larger than 1.
GPN does have higher angle errors than RPN when the aspect ratio is close to 1 as
expected. However, GPN achieves slightly lower angle errors than RPN when the aspect ratio
is larger than 2, even GPN is not directly regressing the rotation angle. The general
trend that the angle errors are lower when the aspect ratios are larger still holds
on the test set for both GPN and RPN (figures are shown in the Supplementary).

%------------------------------------------------------------------------
\section{Discussion}
In this work, we present Gaussian Proposal Networks (GPNs), a new extension to
the popular Region Proposal Networks (RPNs)~\cite{ren2015faster}, to detect 
bounding ellipses of lesions on CT scans. Compared to RPN that uses multiple
regression losses for bounding box localization,
GPN views bounding ellipses as 2D Gaussian distributions on the image plane, and
optimizes the KL divergence between the proposed Gaussian and the ground truth
Gaussian for localization. We show the KL divergence loss is closely connected to
the regression loss used in RPN. On the large-scale medical imaging dataset DeepLesion~\cite{yan2018deeplesion},
GPN significantly outperforms RPN for bounding ellipse detection across different
experimental settings thanks to much lower localization error through the KL divergence
loss. Further error analysis reveals that directly regressing the ellipse rotation
angle may be unnecessary when the ellipse aspect ratio is close to 1.

We intend to further investigate the efficacy of GPN on nature image datasets
with bounding ellipse annotations, such as the FDDB benchmark~\cite{jain2010fddb}.
It is also interesting to comprehensively test the KL divergence loss derived
in Section~\ref{sec1.3} for general bounding box localization on PASCAL VOC~\cite{everingham2010pascal}
and MS COCO~\cite{lin2014microsoft} and see if it also outperforms the current
regression loss in RPN.

%\section*{References}

\small

\bibliographystyle{abbrv}
%\bibliographystyle{unsrt}
%\bibliography{document} %mbe_report

\end{document}